\documentclass{ifacconf}

\usepackage{graphicx}      
\usepackage{natbib}        
\usepackage{leftindex}
\usepackage{amsfonts}
\usepackage{float}
\usepackage{soul}
\usepackage{xcolor}
\usepackage{tabularx}
\graphicspath{{Images/}} 

\begin{document}
\begin{frontmatter}

\title{A Hierarchical Control Architecture \\ for Space Robots in On-Orbit \\ Servicing Operations} 

\author[First]{Pietro Bruschi}

\address[First]{Dipartimento di Scienze e Tecnologie Aerospaziali, Politecnico di Milano,
Via La Masa 34, 20156 Milano, Italy (e-mail: \{pietro.bruschi\}@polimi.it).}

\begin{abstract}                
In-Orbit Servicing and Active Debris Removal require advanced robotic capabilities for capturing and detumbling uncooperative targets. This work presents a hierarchical control framework for autonomous robotic capture of tumbling objects in space. A simulation environment is developed, incorporating sloshing dynamics of the chaser, a rarely studied effect in space robotics. The proposed controller combines an inner Lyapunov-based robust control loop for multi-body dynamics with an outer loop addressing an extended inverse kinematics problem. Simulation results show improved robustness and adaptability compared to existing control schemes.
\end{abstract}


\end{frontmatter}

\section{Introduction}
\label{sec:introduction}
The growth of space debris in Low Earth Orbit (LEO) poses a critical risk to the sustainable use of space. The Kessler syndrome describes the self-sustaining cascade of collisions that could render orbital regions unusable (see \cite{kessler_collision_1978}). To mitigate this threat, two key strategies have emerged: Active Debris Removal (ADR) and In-Orbit Servicing (IOS). ADR focuses on the active removal of defunct satellites and fragments, while IOS extends the operational lifetime of active satellites through tasks such as refueling, repair, and upgrading, as explained in \cite{flores-abad_review_2014, shan2016}.

Space robots represent a promising solution for both ADR and IOS. The design of a coordinated controller for this kind of systems, requiring autonomous capabilities in space environment, is complex due to the dynamic couplings between the spacecraft and the robotic arm. For this reason, they have been studied for many years, starting from the pioneering work of \cite{papadopoulos_coordinated_1991} up to the most recent works of \cite{giordano_coordination_2020} and \cite{giordano_coordinated_2019}. 

The inherent complexity of robotic system is also due to the presence of uncertainties and external disturbances, which can be mitigated using robust control techniques. The works of \cite{dubanchet_modeling_2015} and \cite{faure_h_2022} represent the state of the art in the context of $H_\infty$ control for space robotics systems.

This manuscript presents research on the development of innovative control schemes tailored for robotic capture in space. Building on prior work in combined controllers for space robots, such as the one of \cite{pavanello_combined_2021}, a hierarchical architecture is proposed where only the end-effector reference is imposed, while base and manipulator motions are coordinated automatically. A distinctive feature of this work is the explicit modeling of propellant sloshing in the chaser spacecraft, integrated directly into the recursive dynamic model. To the best of the author’s knowledge, this formulation is novel in the space robotics literature.

The proposed controller relies on a hierarchical architecture (see \cite{invernizzi_hierarchical_2020}) and consists of two nested loops. The inner loop employs a Lyapunov-based robust design with conditional integrators, following \cite{invernizzi_global_2022} and \cite{burger_disturbance_2011} to ensure stability and robustness against uncertainties. The outer loop addresses the inverse kinematics problem for space robots in a novel way, extending traditional formulations to include base motion, and hybrid hysteretic quaternion-based attitude regulation, to avoid unwinding as proposed in \cite{mayhew_quaternion-based_2011}. This structure allows for wider and more flexible maneuvers than previous approaches.

The performance of the controller is assessed in a representative IOS mission scenario involving capture of a semi-cooperative, tumbling satellite in LEO. Such a scenario is taken from \cite{pavanello_combined_2021}, scenario 2. Simulations are conducted under two distinct initial conditions (close and far capture), and results are compared against a reference controller. The outcomes confirm improved robustness, adaptability to unfavorable conditions, and effectiveness in managing sloshing effects Furthermore, the new controller effectively avoids actuators saturation.

The remainder of this manuscript is organized as follows: Section 2 describes the dynamic modeling, including the sloshing formulation; Section 3 details the hierarchical control design; Section 4 presents simulation results; finally, Section 5 summarizes conclusions and suggests directions for future research.

\section{Dynamics Formulation}
\label{sec:dynamics}
The dynamics of space robots, expressed in joint-space formulation, can be described making use of a recursive approach, called Newton-Euler formulation, or with a closed-form approach, namely the Euler-Lagrange formulation (see \cite{siciliano_springer_2008}). Both formulations have been developed for different reasons: the former is used to simulate the dynamics of the system, while the latter is exploited to reconstruct the dynamic matrices inside the controllers. In the recursive formulation, the sloshing phenomenon inside the tank has been added, as described in Section \ref{sec:sloshing}. \\
In the Euler-Lagrange formulation, the state $x$ is composed of 13 variables (supposing a minimal representation of the base orientation), expressing the pose (namely the the orientation $\eta_0$ and the position $p_0$) of the base and the joint angles $q$ of the arm's configuration, while the state derivative $\tilde{v}$ includes the base twist ($\nu_0$) and the joint rates ($\omega_q$):
\begin{equation}
\nonumber
    x = \begin{bmatrix}
        \eta_0 \\
        p_0    \\
        q
        \end{bmatrix}, \quad     \tilde{v} = \begin{bmatrix}
                    \nu_0 \\
                    \omega_q
                    \end{bmatrix}.
\end{equation}
Hence, the equations of motion are expressed as: 
\begin{equation}
\label{eq:eq1}
    M_{CF} \left( x \right) \dot{\tilde{v}} + C_{CF} \left( x, \tilde{v} \right) \tilde{v} = u
\end{equation}
where $M_{CF}$ is the inertia and mass matrix in the closed-form approach, $C_{CF}$ is the matrix accounting for Coriolis and centrifugal effects in the closed-form approach and $u$ is the control action provided by the actuators. \\
In the recursive model, the state is augmented to 15 variables, accounting for additional DOFs provided by the sloshing modeling, supposing to have one tank:
\begin{equation}
\nonumber
    x_{aug} = \begin{bmatrix}
        \eta_0 \\
        p_0    \\
        q      \\
        q_p
        \end{bmatrix}, \quad     \tilde{v}_{aug} = \begin{bmatrix}
                    \nu_0 \\
                    \omega_q \\
                    \dot{q}_p
                    \end{bmatrix}.
\end{equation}
The equations of motion become:
\begin{equation}
\label{eq:eq2}
    H_{RM} \left( x_{aug} \right) \dot{\tilde{v}}_{aug} + C_{RM}\left( x_{aug}, \tilde{v}_{aug} \right) = u_{RM}
\end{equation}
where $H_{RM}$ is the inertia and mass matrix in the recursive approach, $C_{RM}$ is the vector accounting for Coriolis and centrifugal effects, in the recursive approach and $u_{RM} = \left[ u^T \ 0 \ 0 \right]^T$ is the control action, augmented with also the null contribution of the uncontrolled sloshing DOFs.

\subsection{Sloshing Modeling}
\label{sec:sloshing}
\begin{figure}[H]
    \centering
    \includegraphics[width=0.3\textwidth]{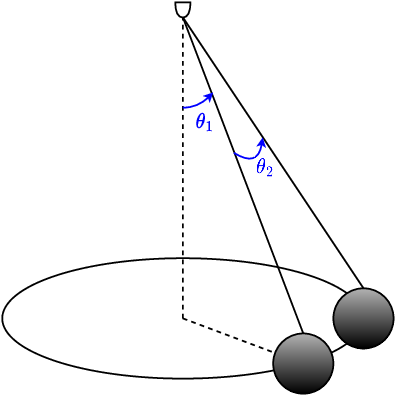}
    \caption{Additional DOFs of the spherical pendulum model.}
    \label{fig:angles}
\end{figure}

Sloshing phenomenon has been included directly in the dynamics of the system, to achieve a more realistic nonlinear simulation, without considering it only in synthesis phase, as done in previous works \cite{pavanello_combined_2021}. An equivalent mechanical model has been selected to represent sloshing. Specifically, a spherical pendulum model allows modeling lateral sloshing behavior. Subsequently, a further analogy between the spherical pendulum and a spherical joint (equivalent to a series of two revolute joints) is exploited, to include the model in the system using the recursive approach mentioned in the previous section, thereby considering the pendulum as an uncontrolled 2 DOFs arm and the whole system as a branched multi-body chain. The graphical representation of the additional DOFs given by the pendulum ($q_p = \left[ \theta_1 \ \theta_2 \right]^T$) is reported in Figure \ref{fig:angles}.\\
Finally, a damping contribution has been added to the model, to achieve a more realistic view. A damping coefficient $\beta_{damp} = 0.0131 \ \mathrm{kg \ m^2 \ s^{-1}}$ is evaluated, based on geometric and fuel properties.

\section{Control Architecture}
\label{sec:control}
\begin{figure}[H]
    \centering
    \includegraphics[width=0.45\textwidth]{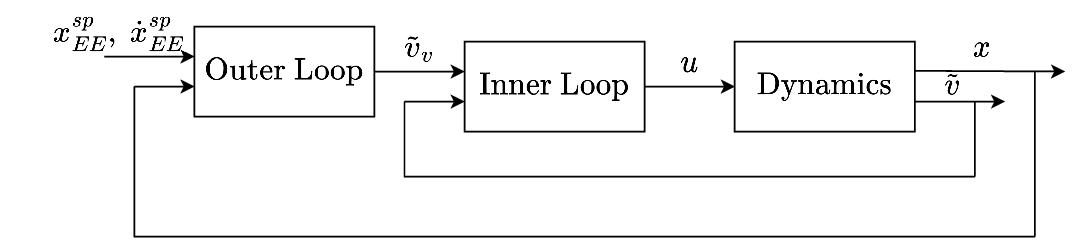}
    \caption{Control system architecture.}
    \label{fig:ctrlsys}
\end{figure}

The control objective is the tracking of a desired trajectory for the end-effector. The states of spacecraft base and manipulator are controlled together and, specifically, a hierarchical architecture is selected, using as a reference the work in \cite{invernizzi_hierarchical_2020}. Hence, the controller is divided into an \textit{Inner Loop}, which tracks a virtual velocity thanks to the computation of the control actions, and an \textit{Outer Loop}, which provides the virtual velocity while controlling the kinematics of the system. A scheme of the complete architecture is shown in Figure \ref{fig:ctrlsys}.
In figure, $x_{EE}^{sp}$ and $\dot{x}_{EE}^{sp}$ indicates, respectively, the desired pose and twist of the end-effector, namely:
\begin{equation}
\nonumber
    x_{EE}^{sp} = \begin{bmatrix}
                    \eta_{EE}^d \\
                    p_{EE}^d
                    \end{bmatrix}, \quad
    \dot{x}_{EE}^{sp} = \begin{bmatrix}
                    \omega_{EE}^d \\
                    v_{EE}^d
                    \end{bmatrix}
\end{equation}
where $\eta_{EE}^d$ is the desired attitude of the end-effector, expressed with quaternions, $p_{EE}^d$ is the desired position and $\omega_{EE}^d$, $v_{EE}^d$ are, respectively, angular and linear velocity setpoints. The current pose of the end-effector depends, instead, upon the entire state. The virtual input is decomposed in the angular velocity ($\omega_v$), linear velocity ($v_v$), and joint rates ($\omega_J$) components:
\begin{equation}
\nonumber
    \tilde{v}_v = \begin{bmatrix}
                  \omega_v \\
                  v_v      \\
                  \omega_J
                  \end{bmatrix}
\end{equation}
while $u$, $x$ and $\tilde{v}$ has been defined above.\\
The design of the two loops is explained with greater detail in the following sections.

\subsection{Inner Loop Design}
The design of the control action that needs to ensure the correct tracking of the virtual input is retrieved leveraging a Lyapunov-based control strategy. A control law that guarantees global stability of the equilibrium point $\tilde{v}_e = 0$, $\tilde{x}_I = 0$ of the system is designed. The Lyapunov function candidate is defined as follows:
\begin{equation}
\label{eq:eq3}
    V = \dfrac{1}{2}\tilde{v}_e^T M_{CF} \tilde{v}_e + \dfrac{1}{2}\tilde{x}_I^T K_I \tilde{x}_I
\end{equation}
where $\tilde{v}_e = \tilde{v}_v - \tilde{v}$ is the error coordinate of the inner loop, $\tilde{x}_I$ is the integral action satisfying $\dot{\tilde{x}}_I = \tilde{v}_e$, and $K_I$ is a positive definite gain matrix to be tuned. \\
The Lie derivative of $V$ along the system error dynamics is studied:
\begin{equation}
\label{eq:eq4}
    \dot{V} = \tilde{v}_e^T M_{CF} \dot{\tilde{v}}_e + \tilde{x}_I^T K_I \dot{\tilde{x}}_I + \dfrac{1}{2}\tilde{v}_e^T \dot{M}_{CF} \tilde{v}_e.
\end{equation}
It can be verified that the control action, which has been defined as:
\begin{equation}
\label{eq:eq5}
    u = K_P \tilde{v}_e + K_I \tilde{x}_I + M_{CF}\dot{\tilde{v}}_v + C_{CF}\tilde{v}_v
\end{equation}
where $K_P$ represent a positive definite gain matrix to be tuned, makes $\dot{V} = -\tilde{v}_e^T K_P \tilde{v}_e \leq 0$. To achieve this result, the property that the matrix $\left( \dot{M}_{CF} - 2 C_{CF} \right)$ is skew symmetric (\cite{siciliano_springer_2008}) has been exploited. \\
The control action is divided into a Proportional-Integral (PI) controller, given by the first two terms, and a feedforward action, provided by the last two terms in the right end side. Therefore, Equation (\ref{eq:eq5}) becomes:
\begin{equation}
\label{eq:eq6}
    u = u_{PI} + u_{FF}.
\end{equation}
The first term has been substituted with a \textit{Conditional Integrator} type controller, following the approach in \cite{invernizzi_global_2022}, to ensure an anti-windup action. Specifically, the controller is defined as:
\begin{equation}
\label{eq:eq7}
\begin{aligned}
& u_{CI} = F_{CI} \mathrm{sat_1} \left( H_{CI}^{-1} \left( G_{CI} \sigma + y \right) \right) \\
& \dot{\sigma} = -G_{CI} \sigma + H_{CI} \mathrm{sat_1} \left( H_{CI}^{-1} \left( G_{CI} \sigma + y \right) \right)
\end{aligned}
\end{equation}
where $\mathrm{sat_1}$ indicates a unit saturation. The matrices $F_{CI}$, $G_{CI}$ and $H_{CI}$ have the same dimensions of $K_P$, $K_I$ and are defined in the following way:
\begin{equation}
\nonumber
\begin{aligned}
    F_{CI} & = \left|\left| K_S \right|\right| \mathcal{I}_{13} \\
    H_{CI} & = K_P^{-1} F_{CI} \\ 
    G_{CI} & = H_{CI} F_{CI}^{-1} K_I
\end{aligned}
\end{equation}
where $K_S$ is a positive definite gain matrix to be tuned and $\mathcal{I}_{n}$ indicates an identity matrix of dimension $n$. \\
The new controller allows, therefore, the switching from a continuous controller (PI) in the unsaturated case to a discontinuous one (sliding mode controller) in the saturated case. Equation (\ref{eq:eq6}) is rewritten as follows:
\begin{equation}
\label{eq:eq8}
    u = u_{CI} + u_{FF}.
\end{equation}
The tuning of the gain matrices has been done leveraging $H_{\infty}$ control and $\mu$-synthesis theory, following \cite{apkarian_multi-model_2014} and \cite{gahinet_structured_2011}, therefore performing a linearization of both the plant and the controller. Specifically, Equation (\ref{eq:eq1}) has been linearized around the capture time of the reference work, obtaining: 
\begin{equation}
\label{eq:eq9}
\begin{cases}
\dot{\tilde{x}} = A_{lin} \tilde{x} + B_{lin} u \\
\tilde{v} = C_{lin} \tilde{x} + D_{lin} u
\end{cases}
\end{equation}
\begin{equation}
\nonumber
\begin{aligned}
& A_{lin} = - \overline{M}_{CF}^{-1} \overline{C}_{CF}, & \quad & B_{lin} = \overline{M}_{CF}^{-1}, \\
& C_{lin} = \mathcal{I}_{13}, & \quad & D_{lin} = 0_{13}
\end{aligned}
\end{equation}
where $\overline{M}_{CF}$ and $\overline{C}_{CF}$ are the dynamic matrices, considered at the specific instant of the capture. \\
The controller is linearized by considering only the first term of Equation (\ref{eq:eq6}) for the tuning of $K_P$ and $K_I$, while only the proportional contribution is retained, for the tuning of $K_S$. \\
The linear system is built considering also the effects of noise and disturbances, as well as simplified models to account for the delay given by sensor and actuators. The requirements are imposed in terms of tracking performances, control effort moderation and disturbance rejection, and the relative weights are constructed. Specifically, the error weight of the sensitivity functions is defined as a diagonal matrix with first order bandpass filters, with cut-off frequencies selected as:
\begin{equation}
\nonumber
    \begin{aligned}
   & \omega_{co_{att}} = & 1.5 \ rad/s \\
   & \omega_{co_{pos}} = & 1 \ rad/s \\
   & \omega_{co_{man}} = & 5 \ rad/s \\
    \end{aligned}.
\end{equation}
referring to the states related to the angular velocities, linear velocities and joint rates. \\
Such weights are imposed, respectively, to the sensitivity function, the control sensitivity and the transfer function from disturbance to error, and are used in the definition of an objective function in the tuning procedure. \\
The controller synthesis is performed and the robustness of the obtained controller is verified. The results are reported in the Table \ref{table:tab1}, where is indicated, for each controller, the $\gamma$-peak value of the objective function, and the lower bound (LB) and upper bound (UB) of the stability margins.
\begin{table}[H]
\centering 
    \begin{tabular}{|p{7em} c c c|}
    \hline
     & \textbf{$\gamma_{peak}$} & \textbf{LB} & \textbf{UB} \\
    \hline \hline
    \textbf{PI-Controller} & 1.07 & 8.5830 & 8.9842  \\
    \hline
    \textbf{P-Controller} & 0.91 & 8.5805 & 8.9843 \\
    \hline
    \end{tabular}
    \\[10pt]
    \caption{Maximum $\gamma$-value and stability margins for the two controllers.}
    \label{table:tab1}
\end{table}

\subsection{Outer Loop Design}
Once the inner loop is solved, a tracking controller, capable to provide the correct virtual input, given the desired coordinates of the end-effector, is developed. The Lyapunov-based approach is considered once again, with the following Lyapunov function candidate:
\begin{equation}
\label{eq:eq10}
    V = \frac{1}{2} | p_{EE}^e |^2 + \frac{1}{2} \mathrm{trace} \left( \mathcal{I}_3 - R_{EE}^e \right)
\end{equation}
which is defined using the error coordinates of the end-effector $R_{EE}^e = \left( R_{EE}^d \right)^T R_{EE}$ (using the attitude matrix for attitude representation) and $p_{EE}^e = p_{EE}- p_{EE}^d$. Note that, the desired values of position and orientation of the end-effector are commanded by the guidance, while the current pose may be retrieved, once the state of the system is available. \\
The Lie derivative of the Lyapunov function is:
\begin{equation}
\label{eq:eq11}
    \dot{V} = \left( p_{EE}^e \right)^T v_{EE}^e + \left( e_{EE}^e \right)^T \omega_{EE}^e
\end{equation}
with $e_{EE}^e = \frac{1}{2} S^{-1} \left( skew \left( R_{EE}^e \right) \right)$. The $S^{-1}\left( \cdot \right)$ map is the inverse of the $S\left( \cdot \right)$ map, defined such that $S \left( y_1 \right) y_2 = y_1 \times y_2$, while the $skew \left( \cdot \right)$ operator is defined such that $skew \left( Y \right) = Y - Y^T$. \\
Therefore, by defining $v_{EE}^e = -K_{pos} p_{EE}^e$ and $\omega_{EE}^e = -K_{att} e_{EE}^e$, with $K_{pos}$ and $K_{att}$ positive definite gain matrices, the stability is guaranteed. Such definitions are used to retrieve the virtual velocities, proposing a new approach for the resolution of the inverse kinematics problem for space robots. The error of the end-effector twist is defined as: 
\begin{equation}
\label{eq:eq12}
\begin{bmatrix} \omega_{EE}^e \\
                 \left( R_{EE} \right)^T v_{EE}^e \end{bmatrix} = J_{v} \tilde{v}_{v} - \left( R_{EE} \right)^T \begin{bmatrix}  R_{EE}^d \omega_{EE}^d \\
                 v_{EE}^d \end{bmatrix}
\end{equation}
where $J_v$ is an extended Jacobian that accounts for the base motion as well. The Jacobian is typically used in robotics to pass from joint rates to end-effector's pose \cite{siciliano_springer_2008}. By inverting the equation, the virtual velocity input is derived: 
\begin{equation}
\label{eq:eq13}
\tilde{v}_{v} = J_v^{\dag} \begin{bmatrix} -K_{att} e_{EE}^e + \left( R_{EE}^e \right)^T \omega_{EE}^d \\
                                           \left( R_{EE} \right)^T \left( -K_{pos} p_{EE}^e + v_{EE}^d \right) \end{bmatrix}.
\end{equation}
The gain matrices are selected as $K_{att} = diag \left( 1, \ 0.99, \ 1.01 \right)$ and $K_{pos} = \mathcal{I}_3$.\\
In Equation (\ref{eq:eq13}), $J_v^{\dag}$ indicates a weighted pseudoinverse, defined as:
\begin{equation}
\label{eq:eq14}
    J_v^{\dag} = W_v J_v^T \left( J_v W_v J_v^T \right)^{-1} 
\end{equation}
where $W_v$ is a state-dependent weight matrix, defined as $W_v = blkdiag \left( \mathcal{I}_3, \ 10 \mathcal{I}_3, \ \epsilon \mathcal{I}_7 \right)$.\\
The $\epsilon$ parameter depends on the relative distance between chaser and target, such that is guaranteed a greater action of the base when the two satellites are far and a greater action of the manipulator when the chaser is approaching the target. Such parameter is defined with the following function:
\begin{equation}
\label{eq:eq15}
    \epsilon \left( \overline{p}_0 \right) = - \frac{90}{\overline{d}} \overline{p}_0 + 100 \frac{\overline{r}_{far}}{\overline{d}} - 10 \frac{\overline{r}_{cl}}{\overline{d}}
\end{equation}
that represent one of the several possible solutions. \\
The variable $\overline{p}_0 = || p_0 ||$ indicates the norm of the position vector of the chaser and, since the target center of mass is centered in the inertial frame, also the relative distance of the two. The term $\overline{r}_{cl}$ is the norm of the closing distance in the guidance algorithm of the reference work \cite{pavanello_combined_2021}, while $\overline{r}_{far}$ must be defined based on the initial conditions of the chaser, such that $\epsilon \geq 0$ is guaranteed at every instant of the capture. For the capture scenarios described in the following section, $\overline{r}_{far} = 2 \overline{r}_{cl}$ is simply defined as twice the closing distance. The term $\overline{d} = \overline{r}_{far} - \overline{r}_{cl}$ indicates their difference.\\
The controller defined in Equation (\ref{eq:eq13}) is further enhanced for what concerning end-effector attitude control, passing to a quaternion representation. A hybrid scheme that allows avoiding the unwinding phenomenon is introduced, following the work in \cite{invernizzi_global_2022}. The controller is designed such that is always pulling in the direction of the shortest rotation. To this aim, a logic variable $h \in \{ -1,1 \}$ that selects the desired rotation is defined, indicating the two possible quaternion representations with $1$ and $-1$. The dynamics of $h$ is defined with the following hybrid system, which allows both a continuous and a discrete evolution of the variable:
\begin{equation}
\label{eq:eq16}
    \begin{cases}
        \dot{h} = 0 & \left( \eta_{EE}^e, h \right) \in \{ h \eta_{s_{EE}}^e \geq -\delta \} \\
        h^+ = -h & \left( \eta_{EE}^e, h \right) \in \{ h \eta_{s_{EE}}^e < -\delta \}
    \end{cases}
\end{equation}
where $\delta = 0.5$ indicates hysteresis half-width of the hybrid solution. \\
The controller is then rewritten as:
\begin{equation}
\label{eq:eq17}
\tilde{v}_{v} = J_v^{\dag} \begin{bmatrix} -K_{att} h \eta_{v_{EE}}^e + \left( R_{EE}^e \right)^T \omega_{EE}^d \\
                                           \left( R_{EE} \right)^T \left( -K_{pos} p_{EE}^e + v_{EE}^d \right) \end{bmatrix}.
\end{equation}
In Equations (\ref{eq:eq16}) and (\ref{eq:eq17}), $\eta_{v_{EE}}^e$ and $\eta_{s_{EE}}^e$ are, respectively, the vector and the scalar part of the quaternion error $\eta_{EE}^e = \left( \eta_{EE}^d \right)^T \otimes \eta_{EE}$.\\
Once the control law that defines the virtual input is derived, the derivative of the latter is evaluated numerically, since it is necessary for the definition of the control law as in Equation (\ref{eq:eq5}). A command shaping filter is used, namely a second order filter that evaluates the virtual acceleration, given as input the virtual velocity. The natural frequency and the damping coefficient of the system are selected as $\omega_{CSF}  = 10 \ \mathrm{rad/s}$ and $\xi_{CSF} = 0.9$. These values are collected in the following matrices:
\begin{equation}
\nonumber
        \Omega_{CSF} = \omega_{CSF} \mathcal{I}_{13}, \quad \Delta_{CSF} = \xi_{CSF} \mathcal{I}_{13}.
\end{equation}
Then, the state-space matrices for the representation of the filter are defined as follows:
\begin{equation}
\nonumber
\begin{aligned}
A_{CSF} & = \begin{bmatrix} 0_{13} & \mathcal{I}_{13} \\
                      -\Omega_{CSF}^2 & -2 \Delta_{CSF} \Omega_{CSF} \end{bmatrix}, \\ 
B_{CSF} & = \begin{bmatrix} 0_{13} \\
                        \Omega_{CSF}^2 \end{bmatrix}, \\
C_{CSF} & = \left[ 0_{13} \quad \mathcal{I}_{13} \right], \\
D_{CSF} & = 0_{13}.
\end{aligned}
\end{equation}
Indeed, the output of the filter is the virtual acceleration $\dot{\tilde{v}}_v$.

\section{Simulation Results}
\label{sec:simulations}
\begin{figure}[H]
    \centering
    \includegraphics[width=0.5\textwidth]{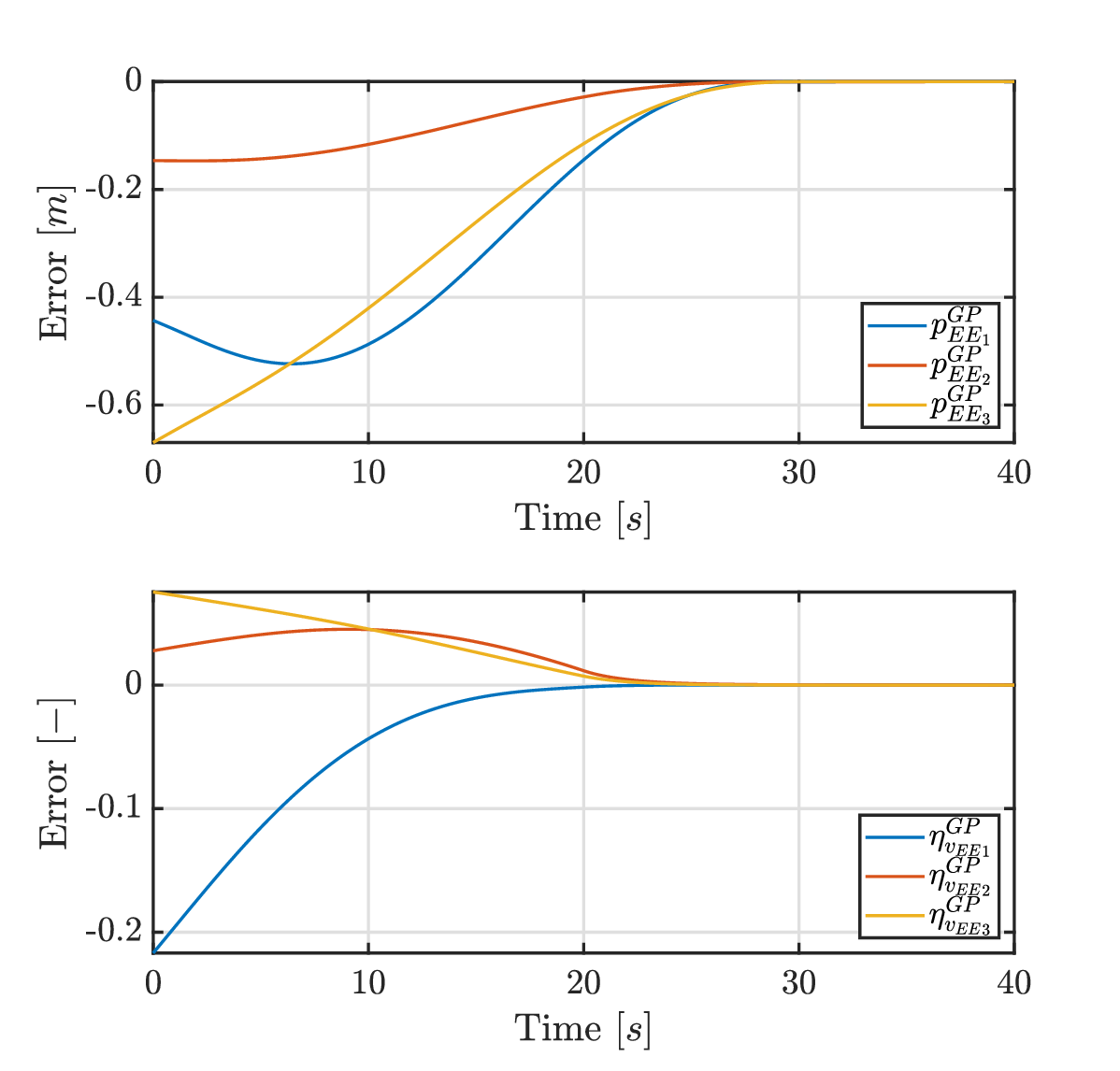}
    \caption{End-Effector/Grasping Point errors - New Controller.}
    \label{fig:ErrNew}
\end{figure}

\begin{figure}[H]
    \centering
    \includegraphics[width=0.5\textwidth]{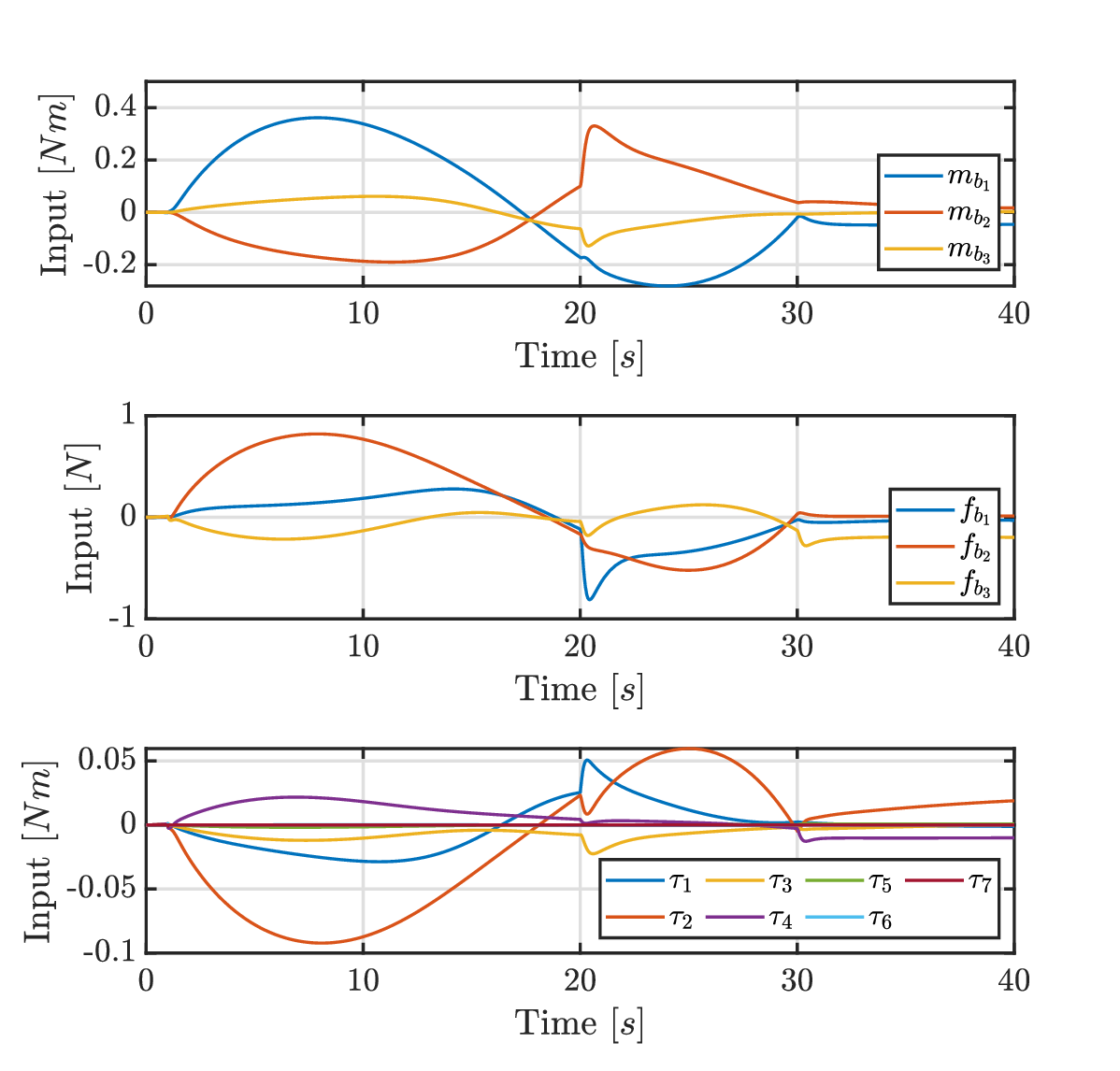}
    \caption{Control input actions - New Controller.}
    \label{fig:CtrlNew}
\end{figure}

\begin{figure}[H]
    \centering
    \includegraphics[width=0.5\textwidth]{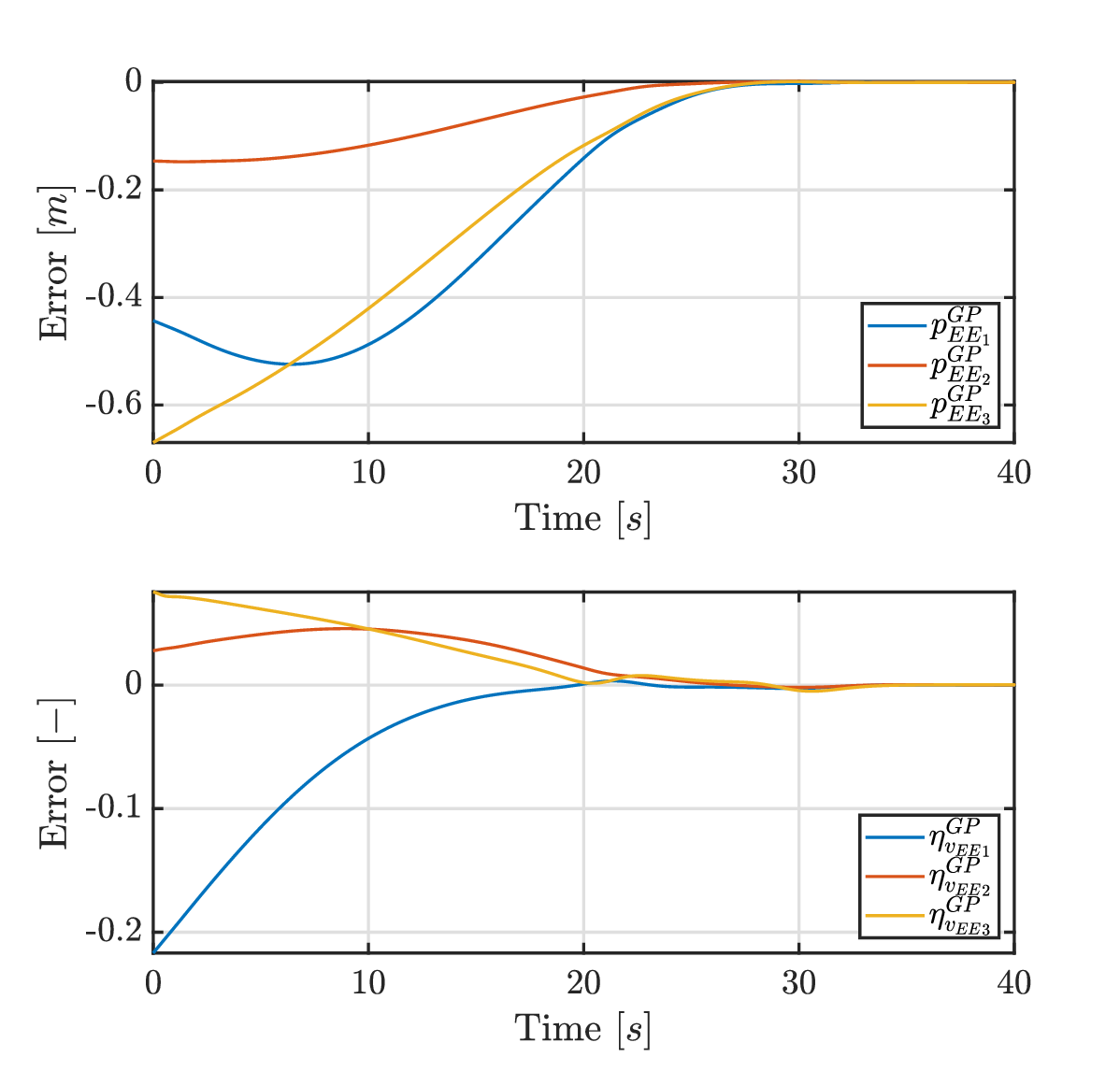}
    \caption{End-Effector/Grasping Point errors - Reference Controller.}
    \label{fig:ErrOld}
\end{figure}

\begin{figure}[H]
    \centering
    \includegraphics[width=0.5\textwidth]{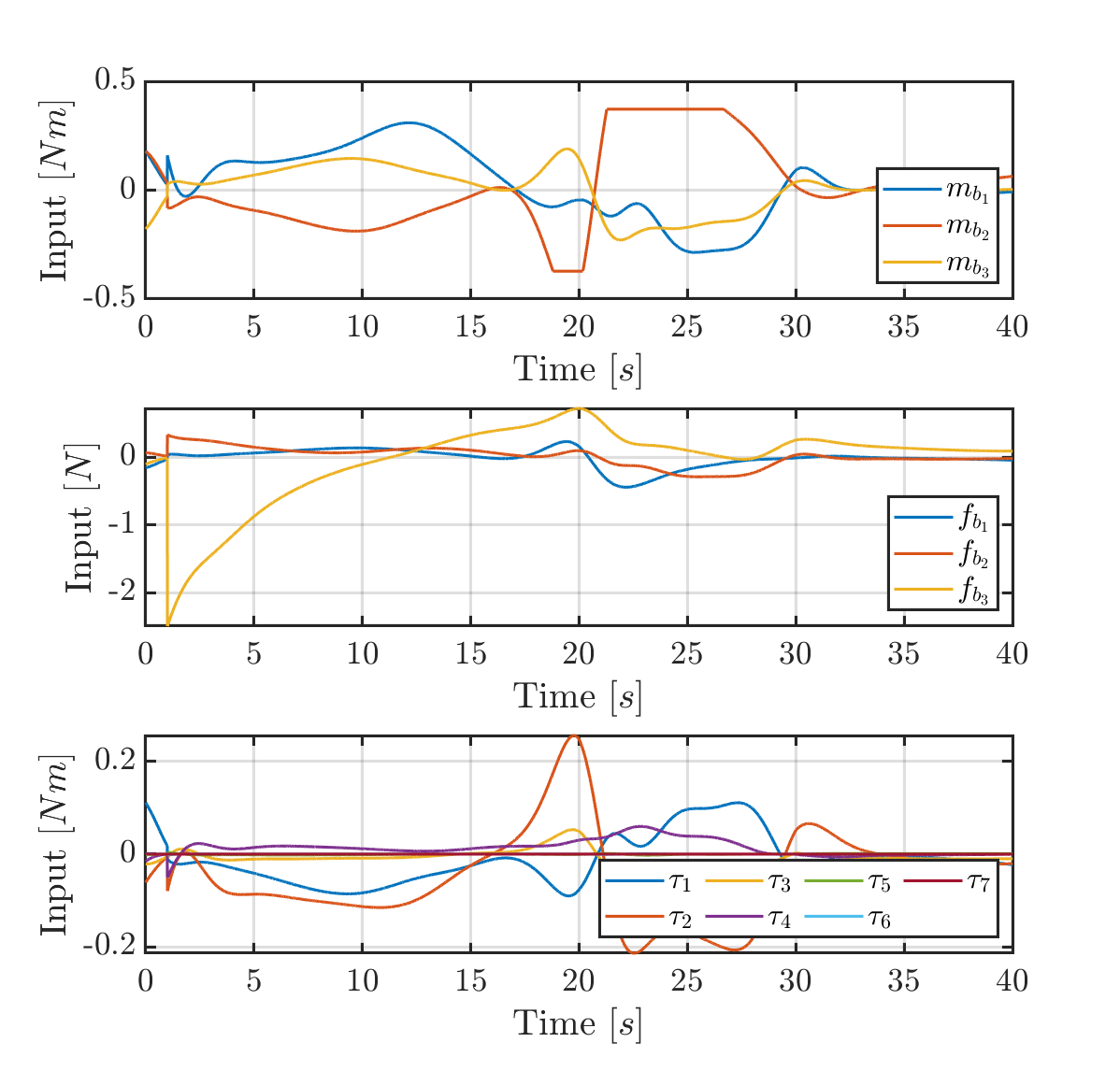}
    \caption{Control input actions - Reference Controller.}
    \label{fig:CtrlOld}
\end{figure}

\begin{figure}[H]
    \centering
    \includegraphics[width=0.5\textwidth]{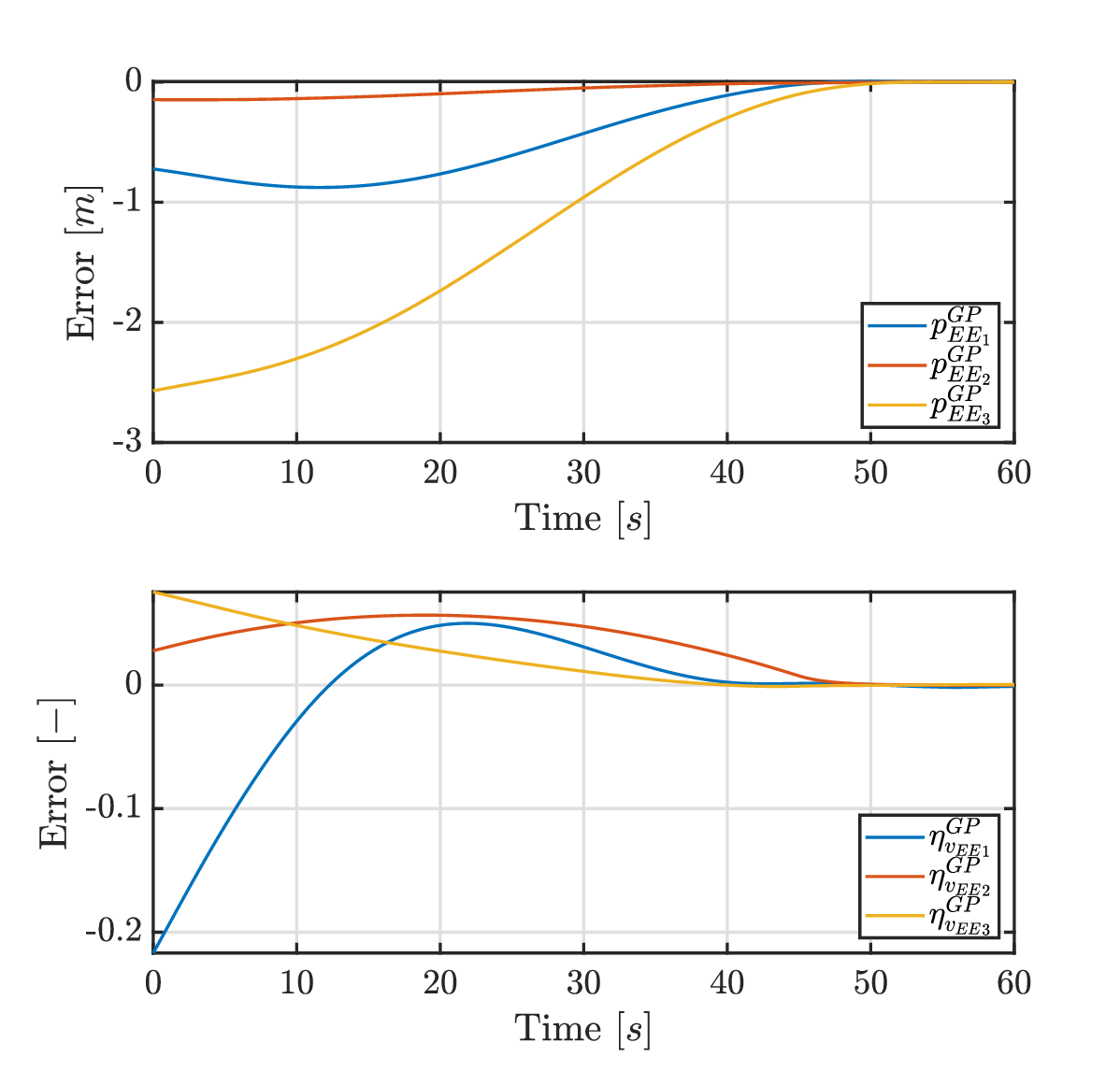}
    \caption{End-Effector/Grasping Point errors - "Far" Capture.}
    \label{fig:ErrFar}
\end{figure}

In this section, the preliminary results obtained in the simulation of the capture scenario are presented. Specifically, the new control architecture is tested and compared to the controller developed in \cite{pavanello_combined_2021} and used as a reference. \\
The guidance algorithm provides the reference trajectory of the end-effector and is characterized by three main instants: at $t_{start} = 1 \ s$ the trajectory generation begins; at $t_{point} = 20 \ s$ the end-effector camera aligns with the grasping point of the target; at $t_{grasp} = 30 \ s$ the position setpoint is reached as well and the capture is completed. End-effector setpoints, for both position and attitude, are generated using a $5^{\mathrm{th}}$ order polynomial to match initial and final conditions. The initial position of the chaser is chosen to be $r_0 = 1.2 \left[-0.35 \ 0 \ -2.375 \right]^T$, which corresponds to a $20 \%$ error with respect to the nominal initial condition of the reference controller. \\
Simulations results show that the new controller is able to track correctly the reference trajectory and performing the capture. The pose error between end-effector and grasping point is reported in Figure \ref{fig:ErrNew}, in terms of position ($p_{EE}^{GP}$) and attitude ($\eta_{v_{EE}}^{GP}$). \\
Moreover, the new controller ensures the avoidance of the actuator saturation, as shown in Figure \ref{fig:CtrlNew}, in which are plotted the control actions of each actuator. Control torques acting on the base are indicated with $m_b$, control forces with $f_b$, and the torques acting on the joints with $\tau$.
The simulation is repeated with the same conditions, but with the reference controller, which is still able to perform the capture.\\
The errors between end-effector and grasping point poses, reported in Figure \ref{fig:ErrOld}, show that the capture execution is less smooth with respect to the new controller.\\ 
Furthermore, the actuator saturation is inevitable in this case for base attitude control, executed using reaction wheels, thus leading to a performance degradation. The control actions are generally higher, as illustrated in Figure \ref{fig:CtrlOld}.
Other capture scenarios have been considered in the thesis. In particular a "Far" capture is studied. The initial position of the chaser has a $100 \%$ error with respect to the nominal initial condition of the reference controller, namely $r_0 = 2 \left[-0.35 \ 0 \ -2.375 \right]^T$.\\
The reference controller was not able to perform the capture, due to the limits of the guidance algorithm, which are overcome thanks to the new control strategy. The new controller can perform the capture, considering a pointing time $t_{point} = 45 \ s$ and a grasping time $t_{grasp} = 55 \ s$ in the guidance algorithm, to account for the greater distance. The errors between end-effector and grasping point poses are reported in Figure \ref{fig:ErrFar}.
\section{Conclusions}
In this work, the problem of controlling with a combined approach  the base and the robotic manipulator of a chaser spacecraft, has been addressed. Specifically, the aim was an automatic partition of the control actions between base and robotic arm actuators. A greater control effort is given either to the base or the manipulator, thanks to the definition of a weighted-pseudoinverse, dependent on the relative distance of the two satellites. A simulation environment has been built up, with the inclusion of the sloshing phenomenon in the dynamics of the system. Then, a new controller has been developed, based on an inner-outer loop paradigm, including a conditional integrator in the inner loop, and an hybrid logic in the outer loop, guaranteeing global tracking of the end-effector state. \\
The efficiency of the new controller has been verified in simulation studies, making a comparison with a reference controller previously developed, showing the better handling of the control allocation for the new controller and its greater adaptability to different initial conditions. In particular, the possibility to perform wider maneuvers is guaranteed thanks to the new control strategy, which does not assign a trajectory on the base, leaving it free to be commanded optimizing the maneuver. \\
Future works may address the subsequent phase of post-capture and target stabilization phase, as well as feasibility studies in case of actuators' failure.

\bibliography{bibliography} 

\end{document}